# ChemBART: Pre-training a Reaction Language Model for Comprehensive Synthesis Planning

Kenan Li [a,1], Yijian Zhang [a,1], Jin Wang [b,1], Haipeng Gan [a], Zeying Sun [b], Xiaoguang Lei [b,*], & Hao Dong [a,*]

[a] State Key Laboratory of Analytical Chemistry for Life Science, Kuang Yaming Honors School, Chemistry and Biomedicine Innovation Centre (ChemBIC), ChemBioMed Interdisciplinary Research Centre at Nanjing University, Engineering Research Centre of Protein and Peptide Medicine of the Ministry of Education, Institute for Brain Sciences, Nanjing University, Nanjing 210023, China.

[b] Beijing National Laboratory for Molecular Sciences, Key Laboratory of Bioorganic Chemistry and Molecular Engineering of the Ministry of Education, College of Chemistry and Molecular Engineering, Peking-Tsinghua Center for Life Sciences, and New Cornerstone Science Laboratory, Peking University, Beijing 100871, China

[1] these authors contributed equally

* Corresponding authors: xglei@pku.edu.cn (X.G.L), donghao@nju.edu.cn (H.D.)

**Abstract:** Recent advances in large language models (LLMs) have demonstrated transformative potential across diverse fields. However, in computer-aided synthesis planning (CASP), current LLM methodologies are often restricted to single-molecule representations, limiting their application to isolated, single-task scenarios such as precursor prediction. Addressing this gap, we present ChemBART, a BART-based language model that adopts a reaction-centric pre-training strategy. By leveraging a mask-filling pre-training task on holistic reaction expressions (Reactant > Reagent > Product), ChemBART effectively solves a variety of chemical problems, including precursor/reagent generation, temperature-yield regression, and molecular property classification. The reaction-centric pre-training further enables the integration of ChemBART with Monte Carlo tree search (MCTS) for multi-step synthesis route design, where the model serves to optimize the policy and value functions within a reinforcement learning framework. Beyond single-molecule pre-trained LLMs that are constrained to specific applications, ChemBART unifies diverse chemical tasks to facilitate synthesis planning with comprehensive implementation information. Most significantly, the proposed multi-step synthesis routes and reaction conditions were validated in the wet lab, confirming the discovery of shorter synthetic pathways with a >20% yield improvement over established literature benchmarks. Our work highlights the potential of reaction-focused pre-training and showcases the broad utility of ChemBART in closing the loop between in silico prediction and wet-lab synthesis.

# Introduction

The field of natural language processing has witnessed remarkable achievements with the advent of large language models (LLMs) based on the Transformer architecture[1]. Prominent examples include GPT[2], BERT[3], and BART[4]. Recently, the simplified molecular input line entry system (SMILES) notation[5], a standard representation for chemical molecules, has been widely seen as a language of the chemical world. Consequently, LLMs have been increasingly applied to acquire chemical knowledge by learning the textual information embedded within SMILES strings.

A particularly successful application based on this concept is precursor prediction, exemplified by models such as AutoSynRoute[6], Molecular Transformer[7], GTA[8], and Graph2SMILES[9], etc. These systems typically input a target product molecule (often as its SMILES string) into an LLM, which then generates the synthetic precursors. This methodology is usually referred to as "template-free," as reaction rules are implicitly embedded within the parameters learned by the LLMs during training.

Notably, the field also features template-based retrosynthesis approaches as a complementary paradigm. Unlike their template-free counterparts, these methods utilize explicitly defined reaction rules to generate precursors. Synthia(Chematica)[10] and AllChemy[11] are well-established methods in this category, which have demonstrated remarkable success in both experimental and computational validations. Their applications span the synthesis of drug-like molecules[12,13], non-trivial natural products[14], as well as in silico validations for complex peptides[15]. These achievements underscore template-based systems as a mature and reliable alternative for retrosynthetic analysis.

Modern LLMs have widely adopted the idea of pre-training – training models to recover a large amount of unlabeled data can boost downstream tasks with only a small amount of labeled data. Some language models based on SMILES notation, such as RXNGraphormer[16], Chemformer[17], X-Mol[18], SMILES-BERT[19], MolBert[20], T5Chem[21], ChemLM[22], CleanMol,[23] and ReactSeq[24], have also conducted similar pre-training on datasets of single molecules. For example, Chemformer[17] undergoes pre-training on mask-filling and synonym transformations of molecule SMILES expressions; T5Chem[21] adopted only mask-filling for single molecules; while RXNGraphormer[16] designed a binary classification on the encoder output of single molecules as pre-training. The pre-training is followed by fine-tuning for downstream tasks, such as forward and retro synthesis prediction, molecular structure optimization, and regression of molecular properties and biological activities.

However, chemical reactions are unquestionably essential in chemical contexts, as they encompass rules governing interactions among atoms and functional groups.

Presumably, pre-training exclusively on single-molecule datasets allows the model to learn SMILES syntax and valency rules, yet it is likely insufficient for acquiring comprehensive knowledge of chemical systems. Moreover, when mask-filling parts of single molecules, diverse substituents and carbon chain structures in molecules create multiple plausible answers, thereby complicating model optimization and hindering the model's ability to learn definitive representations. The limitations of LLMs pre-trained on single molecules and the success of LLMs in precursor prediction thus inspire us to move pre-training forward to chemical reactions.

PMSR[25] introduced a pre-training approach incorporating reaction category information. While offering insights into chemical knowledge from reactions, the specificity of the pre-training objective may limit generalization. Evaluation is currently limited to precursor prediction, leaving broader applicability untested.

Recently, RSGPT[26], a retrosynthesis model built upon LLaMA2, built a training set of over 1 billion mock reactions generated by a template-based method. Using pre-training, which generates entire reactions from the beginning, it achieves a state-of-the-art (SOTA) top-1 accuracy of 63.4% with 3.2 billion parameters, significantly outperforming prior models.

Some works in the literature have attempted to extend retrosynthesis tasks beyond single precursor prediction. For example, the Triple Transformers Loop (TTL)[27] model employs separate Transformer models to predict precursors, reagents and products separately, with product prediction validating the others. This approach provided an important reference for our research; however, it required training distinct models for each task.

Our goal is to leverage reaction-level pre-training to facilitate simultaneous multi-task learning using a single model. While recent works like RSGPT[26] and Chemformer[17] have advanced reaction representation, our implementation of ChemBART specifically focuses on integrating diverse tasks such as precursor prediction, yield regression, and MCTS planning within a single fine-tuning framework. Unlike prior pre-trained LLMs on single molecule SMILES, which were constrained to predicting individual tasks separately, ChemBART can handle diverse chemical tasks and integrate them for full synthesis planning (**Fig. 1**). Specifically, we proposed a novel pre-training strategy via mask-filling of precursors, reagents, and products in reactions, enabling single-step reaction predictions. The reaction-centric pre-training further leads to performance comparable to SOTA methods in various downstream fine-tuning tasks for reaction-relevant data, including temperature, yield, and molecular properties like biochemical activity, aiding synthetic path planning. Even more intriguingly, our pre-training strategy enables a new model capability to fit the policy and value functions in Monte Carlo tree search (MCTS)[28] through first-stage supervised fine-tuning followed by second-stage reinforcement learning under our ReSynZ training framework[29], which prior molecule-based pre-trained models exhibit almost no capability to acquire, thereby directly enabling multi-step synthetic route planning with self-evolving global optimization.

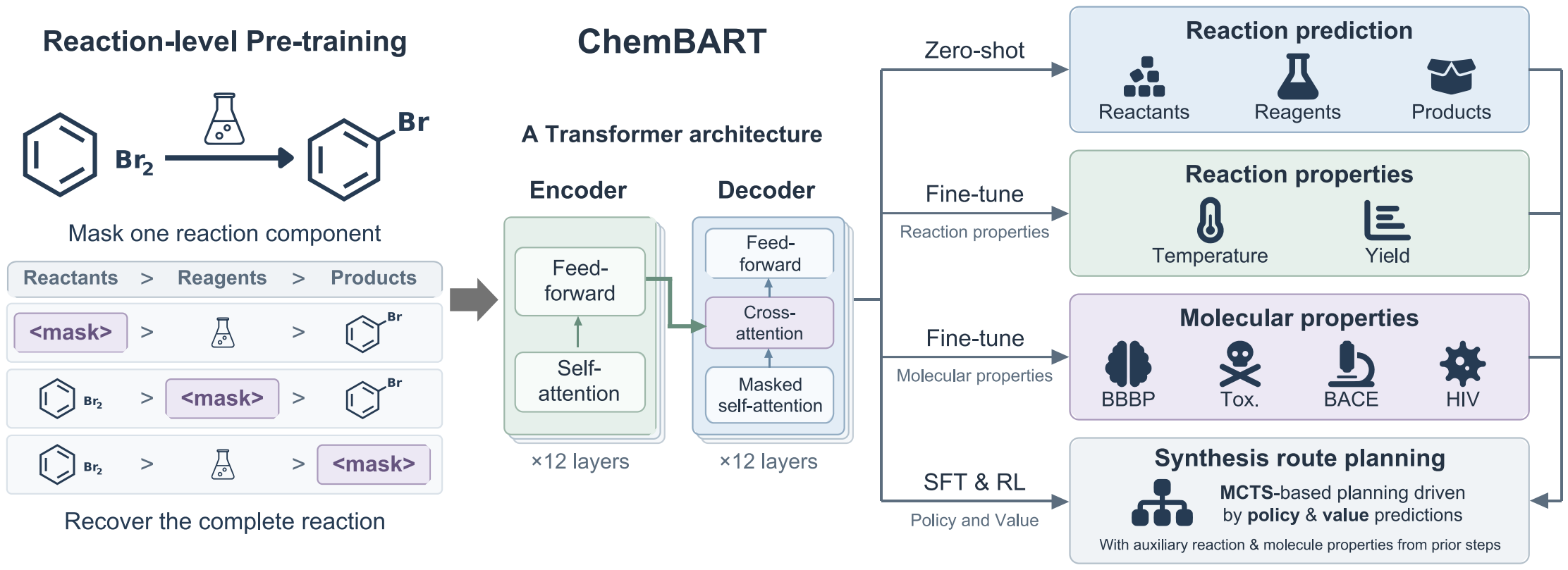

**Fig. 1 | Overview of ChemBART.** ChemBART is a language model pre-trained on chemical reaction data via a mask-filling objective. The model enables direct prediction of reaction precursors and reagents using its pre-trained parameters. Through fine-tuning, ChemBART extends its utility to detailed reaction analysis and molecular property prediction, streamlining chemical synthesis. Through policy and value function learning via supervised fine-tuning (SFT) and reinforcement learning (RL), the model drives Monte Carlo Tree Search (MCTS) to enable multi-step synthesis planning, leveraging insights gained from auxiliary tasks. These computational routes guide wet-lab validation, which in turn provides experimental feedback to refine the model.

After 11 rounds of pre-training (Supplementary **Figs. S1-S2**), ChemBART achieved convergence, resulting in 88.7% top-10 accuracy in precursor prediction and 85.8% in reagent prediction, comparable to the state-of-the-art RSGPT[26] model while using only one tenth of model parameters. For reaction data, it had a 25℃ mean absolute error (MAE) for temperature, equivalent to an 8% MAE relative to the temperature range of -110℃ to 210℃, and an 18% MAE for yield across categories, with an 8% MAE for yield on the Suzuki-Miyaura cross-coupling dataset[30]. In biochemical property prediction, ChemBART achieved SOTA performance on two of five tasks. For MCTS parameter fitting, ChemBART outperformed traditional “template-based” networks. Perhaps most compellingly, ChemBART-designed multi-step synthesis routes and reaction conditions were validated experimentally, confirming shorter pathways with over 20% yield improvement compared to established literature benchmarks. These results highlight ChemBART’s effectiveness and potential utility.

# Results

## 1. Pre-training and fine-tuning workflow

ChemBART is a BART-based language model pre-trained on chemical reactions,

each represented as a SMILES sentence (“reactant > reagent > product”). As illustrated in Supplementary **Tables. S1-S3**, tokenization is performed at the atom/mapping index level, with one of the three components masked at input and reconstructed at output during training. This mask-filling task captures SMILES syntax and chemical principles, providing versatile representations for diverse chemical tasks. We utilized USPTO-full[31] and USPTO-MIT[32] as the pre-training data source. Hereafter, we denote ChemBART pre-trained on USPTO-full as ChemBART-F, the one pre-trained on USPTO-MIT as ChemBART-M, and the randomly initialized model as ChemBART-R. Additionally, to serve as a control for molecule-only pre-training, we also prepare a variant named ChemBART-Mol. This model was pre-trained using the same BART architecture but on a dataset consisting of all the single molecule SMILES strings (excluding reaction context) from USPTO-MIT.

After pre-training, ChemBART is fine-tuned for multiple downstream tasks using task-specific tokens and heads, supporting reaction temperature and yield regression, molecular property classification, and MCTS parameter estimation. The integrated model enables multi-step retrosynthetic route planning and associated reaction implementation information prediction, offering a unified framework for a broad range of chemistry applications.

## 2. Interpretability of ChemBART.

After pre-training, ChemBART embeds chemical knowledge within its parameters. To illuminate its learning process, we analyze the embedding vectors and attention matrices, revealing connections between LLM computation representations and fundamental chemical properties.

**Interpretability of token vector.** To interpret the chemical meaning encoded in the token vectors from ChemBART’s embedding layer, we calculated the Euclidean distances between vectors representing specific element tokens (Cl, Br, O, N, C, Na, K).

The model captured chemical concepts including electronegativity and bond covalency within its embeddings, where smaller distances (darker colors in **Fig. 2a**) denote higher similarity. Intra-group pairs (Cl/Br: 1.02, Na/K: 0.93) were more similar than cross-group pairs (e.g., Cl/Na: 1.07), reflecting electronegativity trends. Similarly, O/Cl similarity (1.00) exceeded O/Br (1.09), consistent with their higher electronegativity and anion basicity. C/O/N formed a distinct cluster reflecting covalent bonding, unlike the ionic-leaning Cl/Br/Na/K group. The lower C/Cl distance versus C/Br further supports stronger C-Cl covalency. A 3D visualization, using two PCA-reduced dimensions and atomic number as the third axis, further indicates that ChemBART learns periodic element trends (Supplementary **Fig. S3**).

Model interpretability was further validated via alcohol oxidation experiments. Dimensionality reduction of decoder end token outputs (**Fig. 2b**) revealed distinct mechanistic patterns, showing consistent vectors for alcohol to aldehyde and aldehyde to acid oxidations. This confirms ChemBART's understanding of SMILES syntax and chemical principles, effectively encoding chemical laws during pre-training.

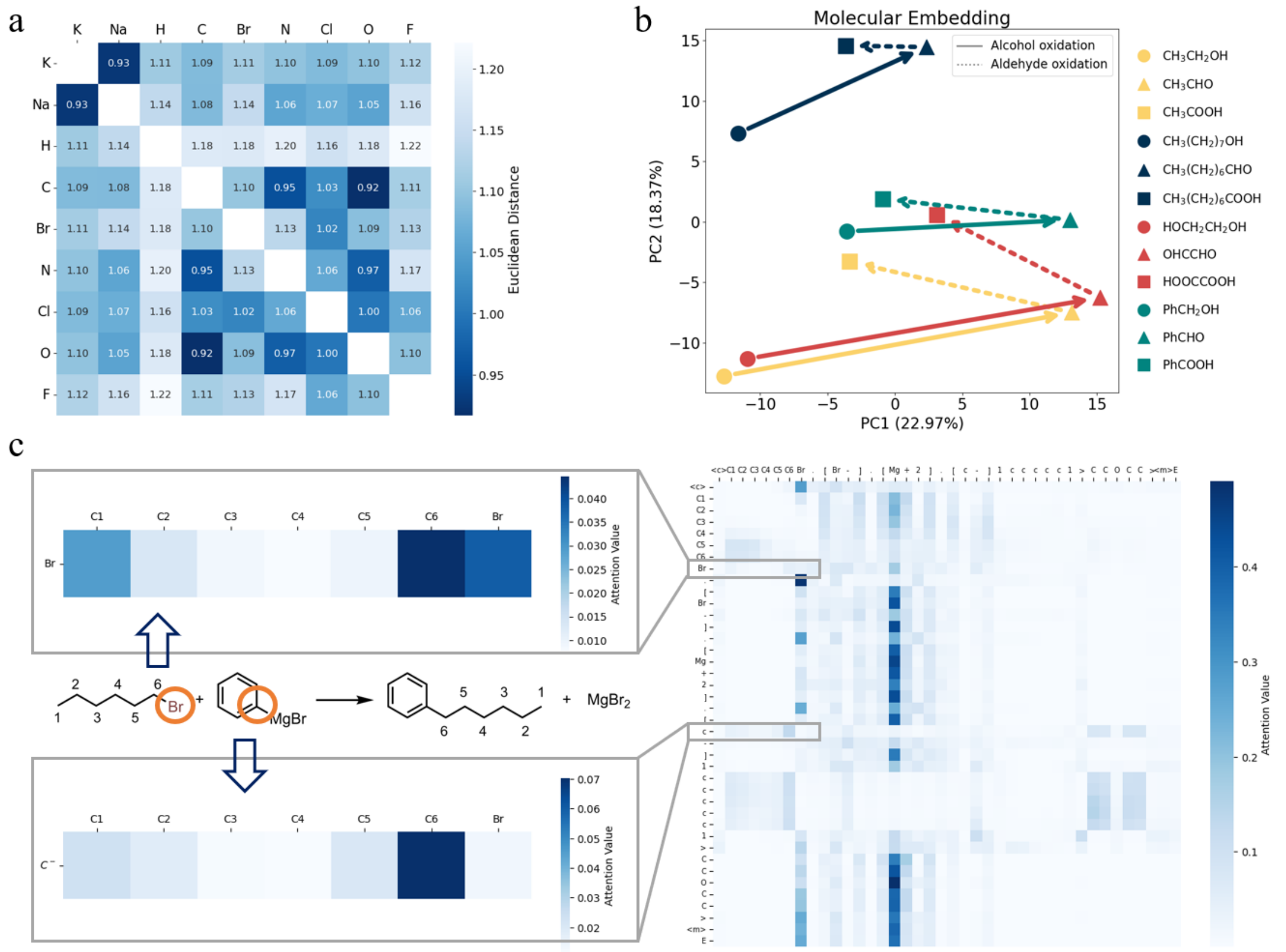

**Fig. 2 | Interpretability analysis of the ChemBART model.** (a) Euclidean distances between characteristic element representations in the embedding layer. (b) 2D projections of the reaction pathway. Alcohols (circle), aldehydes (triangle), and acids (square) of the same series are represented by the same color; arrows indicate oxidation pathways. (c) The encoder's last-layer attention matrix during masked product prediction for a Grignard reaction.

**Interpretability of the attention matrix.** We further investigated ChemBART's reaction understanding using attention mechanisms, taking a substitution reaction between a Grignard reactant **1** and a bromo compound **2**. After masking the product for prediction, the last-layer attention matrix of the encoder (**Fig. 2c**) reveals a strong $C^{-}\rightarrow$C6(-Br) interaction in the alkyl bromide, indicating recognition of the reaction center. Enhanced $C^{-}\rightarrow Mg^{2+}$ attention reflects the model's recognition of charge attraction. Attention decay along the alkyl chain originating from Br atom suggests diminishing reactivity. In contrast, attention of $C^{-}$ to C atoms in the carbon chain increased with chain length from C4 to C1, possibly representing steric effects. Strong local token attention, particularly diagonal ones, indicates that ChemBART recognizes the influences of neighboring atomic groups, which was also seen in decoder and cross-attention (Supplementary **Figs. S4-S5**). Although the mechanism of Grignard reactions remains debated[33], with several possible pathways (such as polar nucleophilic substitution[34-37], single-electron transfer radical mechanism[38,39] and transition-metal-mediated cross-coupling[40,41]), the strong attention between the nucleophilic carbon ($Ph^-$) and the Br-bearing carbon, coupled with $Mg^{2+}$ interaction, aligns most closely with the polar nucleophilic substitution ($S_N2$) category. This shows a learned representation where bond formation is localized between nucleophile and electrophilic carbon, consistent with classical substitution theory. The observed steric gradient in attention (increasing hindrance from C4 to C1) is also in agreement with studies on the steric effect of substituent size on Grignard reaction[34,35,37,42]. Another example of attention matrix in benzene chlorination was shown in Supplementary **Figs. S6-S7**, further illustrating the model's recognition of critical atoms in a reaction. These statistical co-occurrence patterns reflect the capacity of ChemBART to recognize chemically intuitive token associations through training on vast reaction data.

## 3. Single-step retrosynthesis prediction.

Our pre-training task of mask-filling reactions directly enables retrosynthesis precursor, reagent and product prediction. We employed a beam search strategy to give multiple predictions. The top-*k* accuracy on retrosynthesis prediction compared with recent works and ChemBART-Mol/R baselines are shown in **Table 1**. Additional top-*k* accuracy on precursor, product and reagent prediction, along with syntax error rate (invalid SMILES in top-*k* predictions) are shown in Supplementary **Tables S4-S6**.

**Table 1 | Single-step Retrosynthesis Accuracy of Different Template-free Models**

| Model | Parameter size (Million) | Top-1 | Top-3 | Top-5 |
|---|---|---|---|---|
| | **USPTO-50k** | | | |
| **ChemBART-M** | **350** | **63.2** | **81.2** | **85.9** |
| ***ChemBART-Mol*** | ***350*** | ***42.5*** | ***54.0*** | ***59.4*** |
| ***ChemBART-R*** | ***350*** | ***36.2*** | ***46.1*** | ***50.6*** |
| T5Chem[21] | 6.67 | 46.5 | 64.4 | 70.5 |
| RXNGraphormer[16] | 20.5 | 51.0 | 69.0 | 74.2 |
| Graph2SMILES[9] | 21.0 | 52.9 | 66.5 | 70.0 |
| RetroFormer[43] | 7000 | 53.2 | 71.1 | 76.6 |
| Chemformer[17] | 230 | 54.3 | - | 62.3 |
| NAG2G[44] | 64.5 | 55.1 | 76.9 | 83.4 |
| RSGPT[45] | 3228 | 63.4 | 84.2 | 89.2 |
| PMSR (8-layer) [25] | 110 | 67.1 | 82.1 | 85.2 |
| | | **USPTO-MIT** | | |
| **ChemBART-M** | **350** | **63.5** | **82.4** | **86.6** |
| ***ChemBART-Mol*** | ***350*** | ***50.8*** | ***65.2*** | ***72.9*** |
| ***ChemBART-R*** | ***350*** | ***42.4*** | ***52.8*** | ***58.5*** |
| AutoSynRoute - character[6] | 44.4 | 54.1 | 71.8 | 76.9 |
| ReactSeq[46] | 17.4 | 58.9 | 80.5 | 86.4 |
| R-SMILES[47] | 49.6 | 60.3 | 78.2 | 83.2 |
| RSGPT[45] | 3228 | 63.9 | 86.1 | 88.7 |
| | | **USPTO-Full** | | |
| **ChemBART-F** | **350** | **60.6** | **74.8** | **79.8** |
| PMSR (8-layer) [25] | 110 | 45.5 | 60.9 | 65.5 |
| RXNGraphormer[16] | 20.5 | 47.4 | 63.0 | 67.4 |
| R-SMILES[47] | 49.6 | 48.9 | 66.6 | 72.0 |
| NAG2G[44] | 64.5 | 49.7 | 64.6 | 69.3 |
| RSGPT[45] | 3228 | 59.2 | 74.2 | 78.2 |

Our model achieves performance comparable to SOTA template-free tools, (e.g., AutoSynRoute[6]), prior SMILES language models pre-trained on single molecules (e.g., Chemformer,[17] T5Chem,[21] PMSR[25] and ReactSeq[24]), and our own baselines (ChemBART-Mol, the model pretrained with the single molecules from USPTO-MIT, and the random initialized ChemBART-R). Though there is still a small gap with the two SOTA models, RSGPT and PMSR, on USPTO-50k and USPTO-MIT, considering these models are dedicated to retrosynthesis prediction while ChemBART only went through initial pre-training and are designed for multi-task generalization, such performance is already satisfactory. Moreover, as RSGPT uses 20× synthetic training data and 10× model size, due to the limited computing resources available, ChemBART achieves a competitive level of performance using only one tenth of the parameters (Supplementary **Fig. S8**). These findings validate the effectiveness of the ChemBART model, offering three key advantages. Its multi-task pre-training concurrently predicts precursors, reagents, and products, enhancing chemical reasoning via implicit task correlation and tripling training data. Unlike precursor-focused template-free models, ChemBART's simplified design improves generalization and allows fine-tuning across various tasks. Low overall syntax error rates, with minimal impact on generation, indicate robust SMILES syntax acquisition. Reagent predictions show near-zero errors, likely due to their simpler structures and higher training frequency.

## 4. ChemBART fine-tuning: predicting reaction features and molecular properties.

For classification and regression tasks, BART[4] adds special task tokens to the input at the end due to its decoder's masked attention; a linear layer then maps each token's output to the target label. This method, leveraging the model's pre-trained capabilities, is believed to capture overall input information. ChemBART follows this practice. We fine-tuned it on reaction temperature/yield prediction and biochemical property classification.

**Prediction of reaction temperature and yield.** We evaluated ChemBART-F, ChemBART-M, and ChemBART-R for predicting reaction temperature and yield on the Open Reaction Database (ORD)[48]. This dataset consists of experimental records selected by experienced researchers, implying that the recorded temperatures are close to optimal and the yields provide meaningful benchmarks for the given reagents, assuming other variables like concentration were optimized. ChemBART-M showed the best individual performance for property prediction (Supplementary **Tables S7-S8**). We therefore fine-tuned ChemBART-M with task-specific tokens for joint temperature and yield regression, selecting the checkpoint with the lowest validation root mean squared error (RMSE) to mitigate overfitting.

On the ORD dataset[48], the model showed reliable temperature prediction (MAE 25°C, or a relative MAE of 8% across -110 to 210°C, **Fig. 3a**), and could also estimate yield to guide reaction selection (MAE 18%, **Fig. 3b**). For the Suzuki-Miyaura yield dataset[30], our model yielded the $R^2$ of 0.816 across 10 random splits (**Fig. 3c &** Supplementary **Table S9**), rivaling the SOTA Yield-BERT model designed explicitly for this dataset (0.81)[49]. For the Buchwald-Hartwig yield dataset[50], our model yielded 4-test average $R^2$ of 0.74, comparable to state-of-art T5Chem[21] (0.79) and Yield-BERT[49] (0.73) (Supplementary **Table S10**).

Our exploration of fine-tuning strategies (Supplementary **Tables S7-S8**) confirmed that appending task tokens at the reaction's end and using a linear layer without sigmoid constraint on the task token output yields optimal performance. For regression, the BART model's encoder-decoder combination outperformed the BERT model's encoder-only approach, likely due to more parameters. Pre-training significantly boosted regression performance, validating our design.

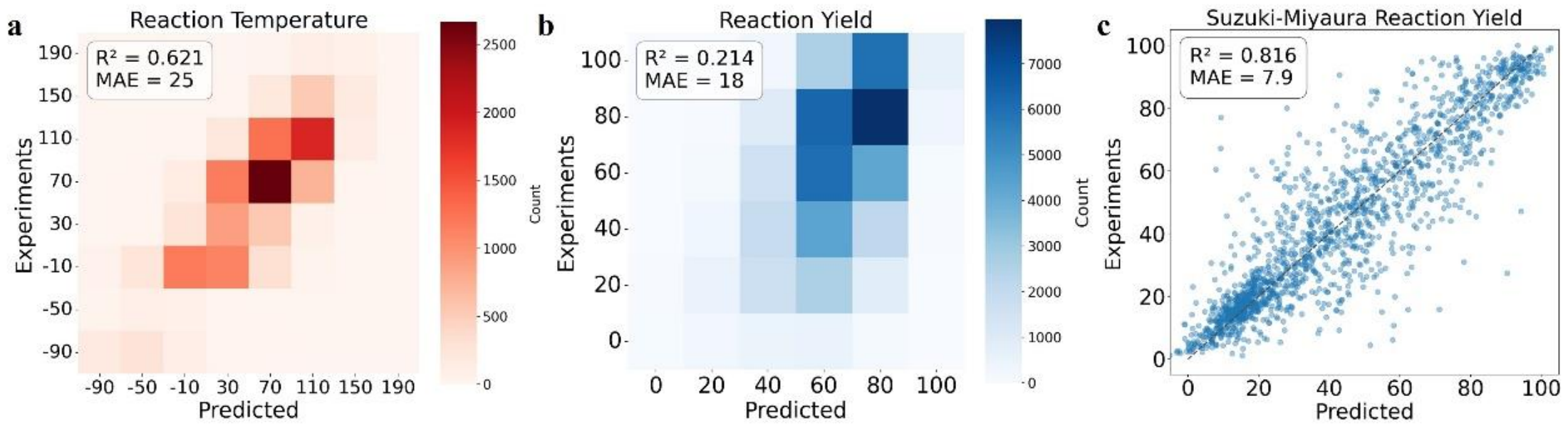

**Fig. 3 | Performance in Temperature and Yield Prediction.** (a) Temperature ($R^2$=0.621, MAE=25°C) and (b) Yield ($R^2$=0.214, MAE=18%) predictions after joint regression of these two factors, with heatmap intensity reflecting data density per 40°C/10% interval. (c) Suzuki-Miyaura yield prediction ($R^2$=0.816, MAE=7.9%).

**Prediction of molecular properties.** For molecular property prediction, we used the ChemBART-M model with an added binary classification layer. We fine-tuned it on five benchmark datasets, BBBP (bioactivity, 2,053), HIV (antiviral, 41,127), BACE (binding affinity, 1,514), Tox21 (toxicology, 8,014), and ClinTox (clinical toxicity, 1,484)[51], using an 8:1:1 split. As shown in **Table 2**, ChemBart achieves SOTA performance on the HIV and Tox21 datasets and comparable performance on the other three datasets when evaluated using the area under the curve (AUC). This suggests that chemical reaction pre-training effectively captures organic molecule interaction principles, validating our paradigm's generalization across biochemical tasks. We also conducted LoRA (Low-Rank Adaptation) fine-tuning [52] in addition to full-parameter fine-tuning, which showed further improvement, likely due to its regularization effect against overfitting. Ablation on ChemBART without pre-training (ChemBART-R) and pre-trained on single molecules from USPTO-MIT dataset (ChemBART-Mol) further demonstrates the effectiveness of pre-training on reaction expressions.

Visualizing pre-training embeddings showed poor separation (Supplementary **Figs. 9a and 9c**), due to the lack of property supervision. Fine-tuning, however, induced apparent clustering (Supplementary **Figs. 9b and 9d**), demonstrating the distillation of task-specific features from the pre-trained model.

**Table 2. AUC index for ChemBART and other models for classification tasks on molecular properties.**

| | **Dataset** | | | | |
|---|---|---|---|---|---|
| | **BBBP** | **HIV** | **BACE** | **Tox21** | **ClinTox** |
| SMILES-Transformers[53] | 0.704 | 0.729 | 0.701 | 0.802 | 0.954 |
| ECFP4[54] | 0.729 | 0.792 | 0.867 | 0.822 | 0.799 |
| GraphConv[55] | 0.690 | 0.763 | 0.793 | 0.829 | 0.807 |
| Weave[56] | 0.671 | 0.703 | 0.806 | 0.820 | 0.832 |
| ChemBERTa[57] | 0.643 | 0.622 | - | 0.728 | 0.733 |
| D-MPNN[58] | 0.708 | 0.752 | - | 0.688 | 0.906 |
| CDDD[59] | 0.761 | 0.753 | 0.833 | - | - |
| MolBERT[20] | 0.762 | 0.783 | 0.866 | - | - |
| GraphCL[60] | 0.685±0.005 | 0.776±0.009 | 0.782±0.012 | 0.754±0.009 | 0.701±0.019 |
| GraphLoG[61] | 0.725±0.008 | 0.778±0.008 | 0.835±0.012 | 0.757±0.005 | 0.767±0.033 |
| Mol2vec[62] | 0.872±0.021 | 0.769±0.021 | 0.862±0.027 | 0.803±0.041 | 0.841±0.062 |
| MolR-GCN[63] | 0.890±0.032 | 0.802±0.024 | **0.882**±0.019 | 0.818±0.023 | 0.916±0.039 |
| MolR-GAT[63] | 0.887±0.026 | 0.794±0.022 | 0.863±0.026 | 0.839±0.039 | 0.908±0.039 |
| MolR-SAGE[63] | 0.879±0.032 | 0.793±0.026 | 0.859±0.029 | 0.811±0.039 | 0.890±0.058 |
| MolR-TAG[63] | 0.895±0.031 | 0.801±0.023 | 0.875±0.023 | 0.820±0.028 | 0.913±0.043 |
| Xmol[18] | **0.962** | 0.798 | 0.872 | - | **0.984** |
| ChemBART-M | 0.903±0.001 | **0.808**±0.015 | 0.873±0.021 | **0.840**±0.011 | 0.872±0.029 |
| ChemBART-LoRA | 0.915±0.012 | 0.789±0.009 | 0.872±0.015 | 0.830±0.014 | 0.920±0.016 |
| ChemBART-Mol | 0.861±0.029 | 0.785±0.026 | 0.850±0.044 | 0.748±0.026 | 0.720±0.051 |
| ChemBART-R | 0.839±0.064 | 0.752±0.005 | 0.780±0.058 | 0.739±0.024 | 0.639±0.091 |

## 5. Synthetic path planning

In previous work, we developed ReSynZ [29], which integrates reinforcement learning and MCTS methods for template-based chemical retrosynthetic analysis. Based on the predictive capability of ChemBART for single-step reaction, we will next leverage it to fit the policy and value functions. By combining these functions with MCTS, we aim to achieve efficient planning of multi-step synthesis pathways.

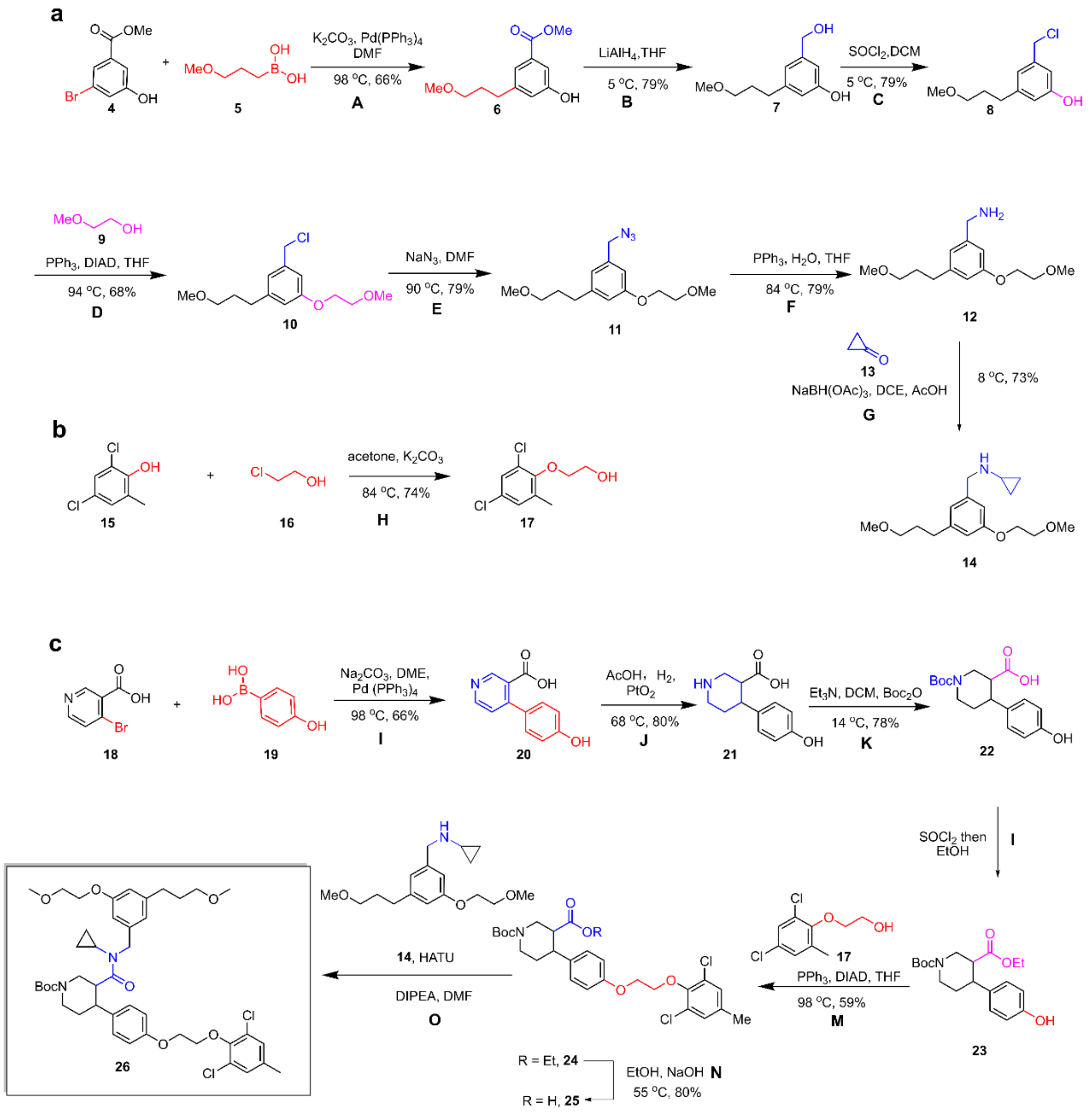

**Fig. 4 | A 15-step synthetic pathway for compound 26[64] designed by ChemBART.** Panels (a), (b), and (c) show the planned paths for intermediates **14** and **17**, and the target molecule **26**, respectively.

Using 12,000 policy and value data points generated by ReSynZ[29] (For further details on data collection, refer to Supplementary **Table S11**), we evaluated three models: ReSynZ[29]'s template-based neural network , which is the dominating policy and value network for multi-step retrosynthesis[65-68] even up until recently[69-73], the pre-trained ChemBART-F/M, and the randomly initialized ChemBART-R. Model performance was quantified by mean RMSE. Training dynamics (Supplementary **Figs. S10-S11**) show ChemBART achieved balanced policy and value prediction, outperforming ReSynZ's poor policy generalization. Seemingly, pre-training enhanced multi-task coordination without overfitting as in the case of template-based networks. Pre-trained ChemBART shows enhanced performance compared to ChemBART-R; even without pre-training, ChemBART-R outperforms template-based neural networks in predicting $p$. Therefore, we saved model parameters only when validation RMSE for both policy and value hit new minima, with final test results in Supplementary **Table S11**.

We then evaluated its predictive performance for multistep synthetic routes with MCTS across various datasets. At the node of the search tree, ChemBART generates top-50 candidates to predict potential precursors and their corresponding reagents. It scores these candidates for chemical feasibility using probabilities from the pre-trained model and synthetic difficulty through the policy and value functions of the fine-tuned model. While ChemBART's single-step prediction accuracy is slightly below the state-of-the-art (**Table 1** & Supplementary **Table S6**), its understanding of reaction logic remains robust for complex, multi-step synthesis planning.

ChemBART-M/F, a template-free method, achieved success rates of 65% and 71% on the Retro*-190 (also known as USPTO-190) benchmark[65] (Supplementary **Table S12**), when using eMolecules, adopted by many established multi-step synthesis works[65,67,74,75], as the starting molecule set.

To the best of our knowledge, the success rate of multi-step synthetic route design via template-free Transformer-based retrosynthesis remains unreported [6-9,16-26,43,44,76-80]. Our work achieves success rates nearly matching template-based methods, alongside consideration for route quality, underscoring the power of ChemBART's pre-training and fine-tuning strategy. **Fig. 4** presents a representative example selected from this dataset. In this case, ChemBART planned a 15-step synthetic route to compound **26** (PubChem CID: 86639854, ID 40 in Supplementary **Table S12**), a designed renin inhibitor[64], with detailed reaction conditions and yields. This example highlights ChemBART's capability for long-path planning, showcasing innovative strategies such as the reductive selectivity (step-J in **Fig. 4**), which could potentially inspire new pathways.

To evaluate predictive performance on contemporary organic molecules, we created a novel multi-step retrosynthesis dataset containing 53 highly diverse and distinct molecules from 30 recent articles in the *Journal of Medicinal Chemistry* in 2025 (denoted as JMC2025 dataset, Supplementary **Table S13 & Figs. S12-S13**). Using this dataset, we investigated different generation strategies: beam search, top-*k* (k=10), and top-*p* (p=0.9) achieved success rates of 88.7%, 84.9%, and 83.0%, respectively (predicted route examples are presented in **Fig. 5**, Supplementary **Figs. S14-S35**). Analysis of predictions on JMC2025 (Supplementary **Fig. S36)** reveals that beam search tends to produce high-confidence but chemically conservative routes, while top-*k* and top-*p* sampling introduce greater synthetic diversity at the cost of increased hallucination risk, highlighting the challenge of balancing novelty and reliability.

We further evaluated ChemBART on 11 novel FDA-approved drugs in 2025[81] that contain stereochemical centers. Default search width is 10, and if the search is blocked at some difficult steps, the width is increased to 50 and then 150. ChemBART reproduced literature routes for all 11 molecules when each step is manually selected (Supplementary **Figs. S37-S47**), demonstrating its application value for drug synthesis. To test upper bounds, we evaluated ChemBART on the Chematica (Synthia) benchmark[12,14]. ChemBART resolved 4 out of 11 molecules with manual step selection (Supplementary **Figs. S48-S51**), indicating a gap between commercial proprietary software like Chematica and open-source models. Notably, automatic MCTS search as introduced above also resolves the 15 molecules, though it results in more difference from literature routes. ChemBART still struggled with ring-closing transformations or highly complex topologies, likely due to the limitations of the public USPTO training set in covering such structures.

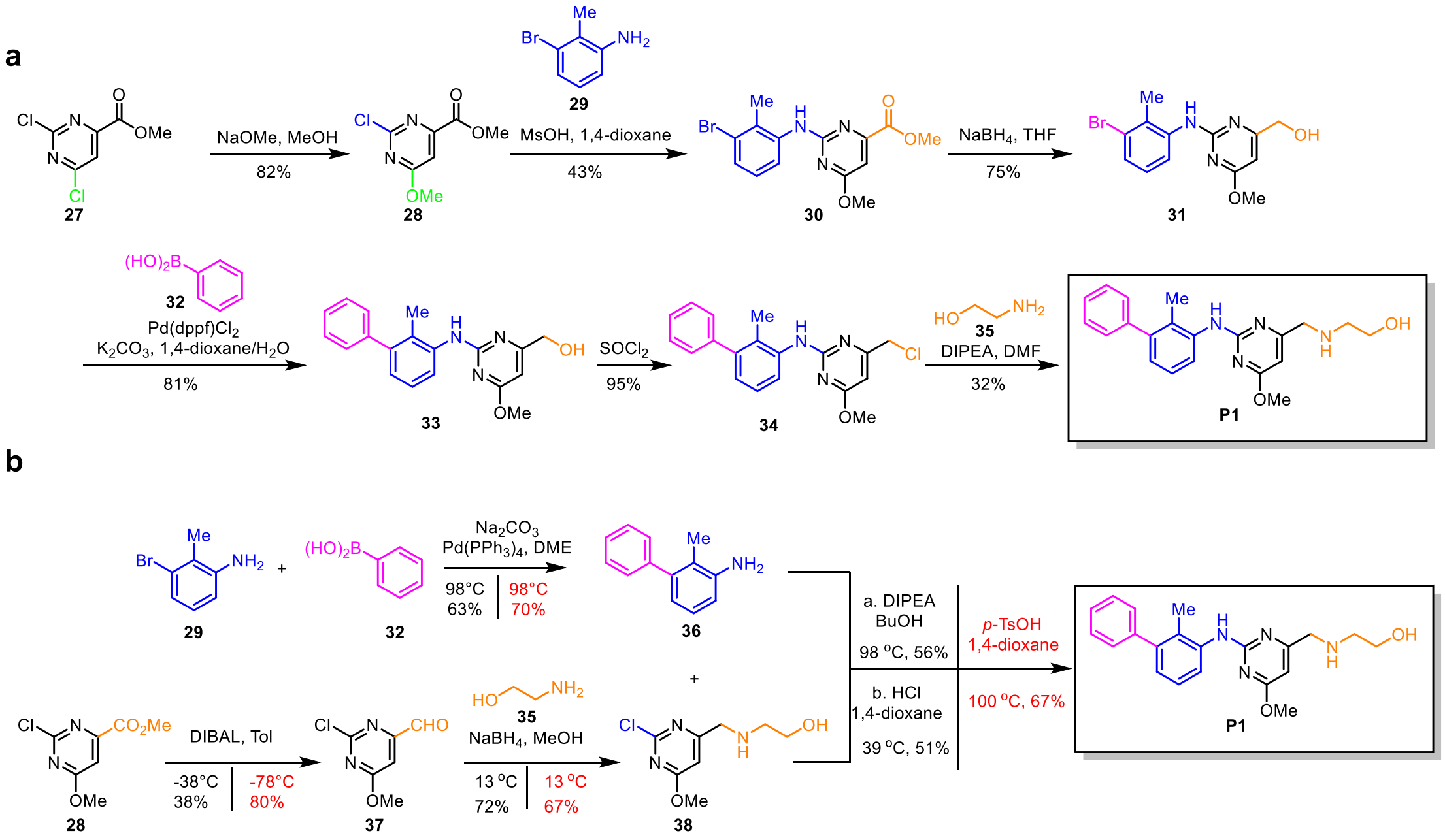

**Fig. 5 | Comparison of synthetic routes for compound P1: Literature-reported vs. ChemBART-predicted and experimentally validated.** (a) The literature-reported experimental route[82]. (b) The ChemBART-predicted synthetic route, including predicted reaction conditions and yields, alongside experimental validation results. Within panel (b), optimized reaction conditions and corresponding yields, determined experimentally, are highlighted in red.

## 6. Wet-lab validation

To further evaluate the applicability of ChemBART in multistep synthetic planning, we selected a representative compound **P1** (ID **3** in Supplementary **Table S13**) from the JMC2025 dataset for wet-lab validation (Supplementary **Section A and Figs. SA1-SA12**). **P1** was designed as a PD-L1/VISTA dual inhibitor and had been previously synthesized in six steps with an overall yield of 6.5% (**Fig. 5a**)[82].

A more convergent route was designed by ChemBART featuring the late-stage $S_NAr$ reaction between amine **36** and alcohol **38** (**Fig. 5b**). Following this route, amine **36** was obtained under Suzuki coupling conditions in 70% yield. Meanwhile, ester **28**, which is common to both the literature[82] and ChemBART-designed routes, was reduced with DIBAL-H to give aldehyde **37**, which subsequently underwent reductive amination to afford alcohol **38**. Finally, the $S_NAr$ reaction between amine **36** and alcohol **38** was achieved under acidic conditions (HCl), albeit with a yield of only 10%. After screening various conditions (see supporting information), we identified *p*-TsOH as the optimal acid, achieving a substitution efficiency of 67%. Overall, the ChemBART-designed route delivered **P1** in an overall yield of 35%, representing an improvement of 27% from ester **28** compared with the original route.

Notably, most reaction conditions proposed by ChemBART were readily applicable in the laboratory, and the incorporation of expert knowledge and further optimization enabled us to achieve the best overall performance, underscoring both the practical utility and the inspirational value of ChemBART's designs.

## Discussion

**The mask-filling of reaction components for pre-training is effective for the model to learn general chemical knowledge.** Various experiments have demonstrated that abundant chemical knowledge captured within the pre-trained parameters significantly enhances performance across various downstream fine-tuning tasks[17-24]. Our analysis reveals that during pre-training, model embedding and attention layers acquire SMILES syntax and chemical knowledge, embedding this directly into parameters and thus improving fine-tuning outcomes on pre-trained models.

However, while recent SMILES LLMs[17-24] often pre-train on individual molecules using tasks like mask-filling or synonym conversion, our results show that this approach (exemplified by ChemBART-Mol) is insufficient for acquiring deep chemical reasoning, as they primarily capture SMILES syntax and fail to deeply embed essential reaction knowledge needed for molecular interconversion. Furthermore, the variability within molecules complicates conventional pre-training tasks (e.g., predicting masked segments), as they lack definitive answers, hindering the learning of consistent chemical rules.

In contrast, the reaction-level pre-training is indispensable for the model to learn the relationships of atoms during a transformation. ChemBART employs a novel three-directional pre-training task with exact answers. This allows the model to learn clear chemical laws, yielding robust pre-training results effectively. Unlike single-molecule LLMs limited to specific tasks, ChemBART can tackle more versatile chemical challenges and integrate these capabilities to accomplish the full synthesis planning cycle, highlighting the advantage of reaction-level pre-training.

**ChemBART achieves high parameter efficiency by utilizing a unified set of**

**weights across multiple tasks, thereby significantly conserving computational resources.** Driven by its pre-training strategy, the model seamlessly handles both generative tasks (such as predicting precursors, reagents, and products), and regression tasks, including temperature and yield optimization, all without requiring task-specific models. Additionally, parameter fitting provides policy and value estimates crucial for synthesis planning. This multi-task capability is particularly economical, as it avoids the need to load and maintain separate models, making efficient use of GPU resources, a critical advantage given the high computational demands of large language models.

While ChemBART is built upon a standard BART-style encoder-decoder architecture and shares conceptual similarities with existing reaction language models, its primary contribution lies in the empirical validation of reaction-level pre-training and the integration of multi-step planning within a unified framework. Through literature review on template-free models[6-9,16-26,43,44,76-80] and multi-step synthesis planning works[65,67,69-75], methods that simultaneously report single-step accuracy and multi-step success rate are rarely reported. Our work contributes to this area by integrating policy and value function learning on a pre-trained language model with explicit success rate metrics for multi-step synthesis planning. This approach effectively balances single-step accuracy with multi-step success rate. It also proposes a practical method to fine-tune a pre-trained language model with first-stage supervised fine-tuning and second-stage reinforcement learning for policy and value prediction, offering insights into policy design for future multi-step synthesis works.

**Generalizability of the reaction-centric pre-training algorithm.** We further trained a T5-base model, with different embedding, attention and fully-connected layer structures and only half of the model size (190M parameters). Results in Supplementary **Table S14** shows the T5 model can fully reproduce the results of ChemBART-M, illustrating the robustness and transferability of our pre-training algorithm to different model structures and sizes.

**Directions for future study.** While our model demonstrates strength in reaction-related tasks, its performance in predicting molecular physicochemical properties, such as activation energy on the QM9 dataset[83], remains limited. This suggests that sequence information alone is insufficient for capturing all relevant aspects of molecular behavior. Potential improvements include integrating 3D structural data via graph neural networks, combining it with the language model output, as done by KFLM2[84], and incorporating sequence attention and structural contact maps, inspired by Uni-Mol[85], to better link sequence and structure representations is another option.

We have also observed efforts to adapt general-purpose LLMs pre-trained on natural languages, such as LLaMA[86] and Qwen[87], using both SMILES strings and chemistry literature[23,84,88-92], or just simply prompt these LLMs using their general knowledge.[93,94] While their advantage over SMILES-only models is yet to be clearly demonstrated, their potential to leverage vast general knowledge and learn from natural language descriptions holds significant promise for enhancing downstream performance and enabling broader application through natural language input.

To further enhance performance, future optimizations will include additional MCTS cycles and iterative retraining. We also aim to implement conditional precursor generation based on reaction types and sites, thereby increasing precursor diversity, tree search flexibility, and the overall success rate of multi-step synthesis planning.

Yield prediction based on the ORD dataset[48] currently shows room for improvement. The training data is inherently noisy because reported conditions are typically experimental estimates rather than theoretical optima. Moreover, predicting yield from a model-derived temperature introduces error propagation. Future work will address this by applying sequential constraints (that is, predicting optimal concentration, then temperature, and finally yield) to improve prediction fidelity.

Finally, while the eMolecules dataset provides a vast library of commercially available compounds, many are economically unfeasible. Currently, when synthesis planning terminates at these expensive intermediates, we dynamically mask them in the dataset to prompt further breakdown by the model. Future directions include incorporating real-time cost analysis to systematically filter the building block library.

# Method

**Definition of chemical sentence and vocabulary list.** We define a complete chemical reaction "reactant > reagent > product" as a sentence in the SMILES[5] language. Then, tokenization and index conversion are required to transform a chemical reaction into a digital representation readable by the language model. In tokenization, the SMILES sentence is segmented into sequences of tokens according to a predefined word list (**Supplementary Tables S1-S3**). In this study, we take a complete atom or atom mapping index as a token. Although AutoSynRoute[6] pointed out after experiments that there is no significant difference in the training effectiveness of taking atoms or separate characters as tokens, we choose to use complete atoms or atom mapping indices as the token units to improve interpretability and reduce computational overhead. Given that the SMARTS[95] language serves as an extension of SMILES, the word list has been designed to accommodate SMARTS syntax as well. This compatibility enables researchers to extend the application of this pre-trained model to address challenges expressed within the SMARTS language framework. In index conversion, each token in the resulting sequence is then mapped to its corresponding index based on the word list.

After passing through the embedding layer of LLM, each token index is transformed into a *d*-dimensional vector, composed of the sum of a lexical vector and a positional vector. In our problem, the lexical vector encodes the chemical semantics of the token. For example, if the token represents a chemical element, its lexical vector captures general chemical properties associated with that element. Although such high-dimensional representations are not directly readable to humans, the vectors corresponding to different tokens are comparable. Specifically, the distance between two representation vectors reflects the similarity of their chemical properties. To quantify this similarity, we used the Euclidean distance *D*(x,y) between the two vectors, $x = [x_1, x_2 ..., x_d]$, $y = [y_1, y_2 ..., y_d]$, where a smaller Euclidean distance indicates greater similarity between the two vectors.

$$D(x, y) = \sqrt{\sum_{i=1}^{d} (x_i - y_i)^2}$$

In our setting, dimension *d* is set to 1024.

**Model structure.** Our language model adopts the framework of the BART-large model,[4] featuring approximately 0.4 billion parameters. Each token is represented by a 1024-dimensional vector, and the model accommodates a maximum input length of 1024 tokens. The attention mechanism comprises 16 attention heads per layer, with the architecture structured into 12 encoder layers and 12 decoder layers. The inclusion of both encoder and decoder components is specifically suited to addressing the diverse requirements in chemical tasks, as the encoder is suited for understanding the semantic meaning, which is the molecular structure in our problem, while the decoder is suited for the generation of molecular structures. Moreover, the decoder can be extended to support classification and regression tasks by integrating specific heads. In addition, the trained encoder can be extracted separately to build a BERT model (referred to as ChemBert in the following context) for dedicated labeling, classification and regression.

As a key component of Transformers,[1] the attention mechanism enables the model to perform far beyond earlier models dealing with sequential problems, such as RNN[96] and LSTM[97]. The basic algorithm of the attention mechanism is that, for the hidden layer representation of a sentence ***X*** input into the attention module,

$$X = [T_1, T_2, ... T_n]^T, \text{where } T_i = [x_1, x_2, ... x_d], x_i \in R$$

where $T_i$ is the representation of token *i* in this hidden layer and *n* is the number of tokens in the sentence, an attention matrix *A* is calculated by

$$K = XM_1$$
$$Q = XM_2$$
$$A = \text{softmax}(\frac{QK^T}{\sqrt{n}})$$

where $M_1$, $M_2$ are the matrices with trainable parameters, and the value of $A_{ij}$ represents the attention of token $i$ to $j$, i.e., the degree of relevance of $j$ for $i$. The representation of token $i$ in the new round output by the module, $T_i'$, is a weighted summation of the vector representations of all the tokens concerning the attention of $i$ to each

$$V = XM_3 = [V_1, V_2, \ldots V_n]^T$$

$$T_i' = \sum_{j=1}^{n} A_{ij} * V_j$$

where $M_3$ is the matrix with trainable parameters, $V_i$ still represents the information of the token $i$, which is linearly transformed on the basis of $T_i$. Therefore, in the next layer, the hidden layer representation of the input sentence $X'$ is

$$X' = AV$$

**Mask-filling pre-training.** In our framework, each "reaction sentence" consists of a triple of reactant, reagent, and product, any of which can be masked. Based on chemical principles, the masked component remains predictable given the remaining parts. Thus, the pre-training task in this study is designed such that an input reaction sentence with one masked component (reactant, reagent, or product) is processed by the decoder to reconstruct the original complete sentence. For example, a reaction with reagents masked is input into the encoder in SMILES:

<cls> c1ccccc1.ClCl > <mask> > Cl-c1ccccc1.HCl <end>

The decoder outputs the whole reaction from scratch:

<cls> c1ccccc1.ClCl > [Fe3+].[Cl-].[Cl-].[Cl-] > Clc1ccccc1.HCl <end>

This approach facilitates the model's acquisition of both fundamental SMILES syntax and essential chemical reaction principles.

The model was pre-trained on the USPTO-full reaction dataset [31] (containing approximately 1.5 million reactions), primarily derived from medicinal chemistry synthesis patents. According to Schneider et al.,[98] the main reaction types in this dataset, ordered by frequency, include: heteroatom alkylation and arylation, acylation, protection/deprotection, C–C bond formation, redox reactions, heterocycle formation, and functional group interconversion or addition. As R-smiles[47] has recently reported that the increase in reaction data quality will significantly increase the performance of language models, we also conducted a pre-training on the USPTO-MIT dataset,[32] a filtered USPTO dataset excluding unreliable reaction records, and compared its performance with the model trained on USPTO-full. A 9:1 split was applied to separate the training and test sets. The USPTO-50k dataset,[98-100], a much smaller data set with 50k pieces of data filtered from USPTO-full, was also used for testing.

The training was performed using the AdamW optimizer with the Binary Cross Entropy loss function, a learning rate of 1e-5, a weight decay value of 1e-4, and a batch size of 256. Pre-training was done on 2*8 Nvidia-A40 GPUs with 1 months.

**Reaction temperature and yield prediction.** We employed our pre-trained model to analyze the temperature and yield data of organic reactions provided by the Open Reaction Database (ORD)[48]. The dataset comprises approximately 700k entries associated with reaction temperatures and yields. We partitioned the dataset into training, validation, and test sets with an 8:1:1 ratio.

To evaluate different fine-tuning approaches, we implemented multiple training schemes. Specifically, we compared the strategy of combining both encoder and decoder for regression tasks (following the BART model strategy) with a strategy that uses only the encoder (aligned with the BERT model strategy). Additionally, we investigated the potential benefits of joint regression across multiple tasks. This involved appending several task-specific tokens at the end of the reaction expression and applying a linear head for each token, to enhance information flow to improve task performance. It should be noted that due to the decoder attention framework, the yield is predicted given the reagent and model-predicted temperature value as a constraint. These approaches are illustrated in **Fig. 6a & 6b**, respectively. Further, we examined whether incorporating a Sigmoid layer to constrain the output range of the regression head could enhance prediction accuracy. We also compared the effectiveness of a Bi-LSTM layer, which integrates the output vectors of all tokens before feeding them into a linear head for regression, against the "one task token for one task" approach.

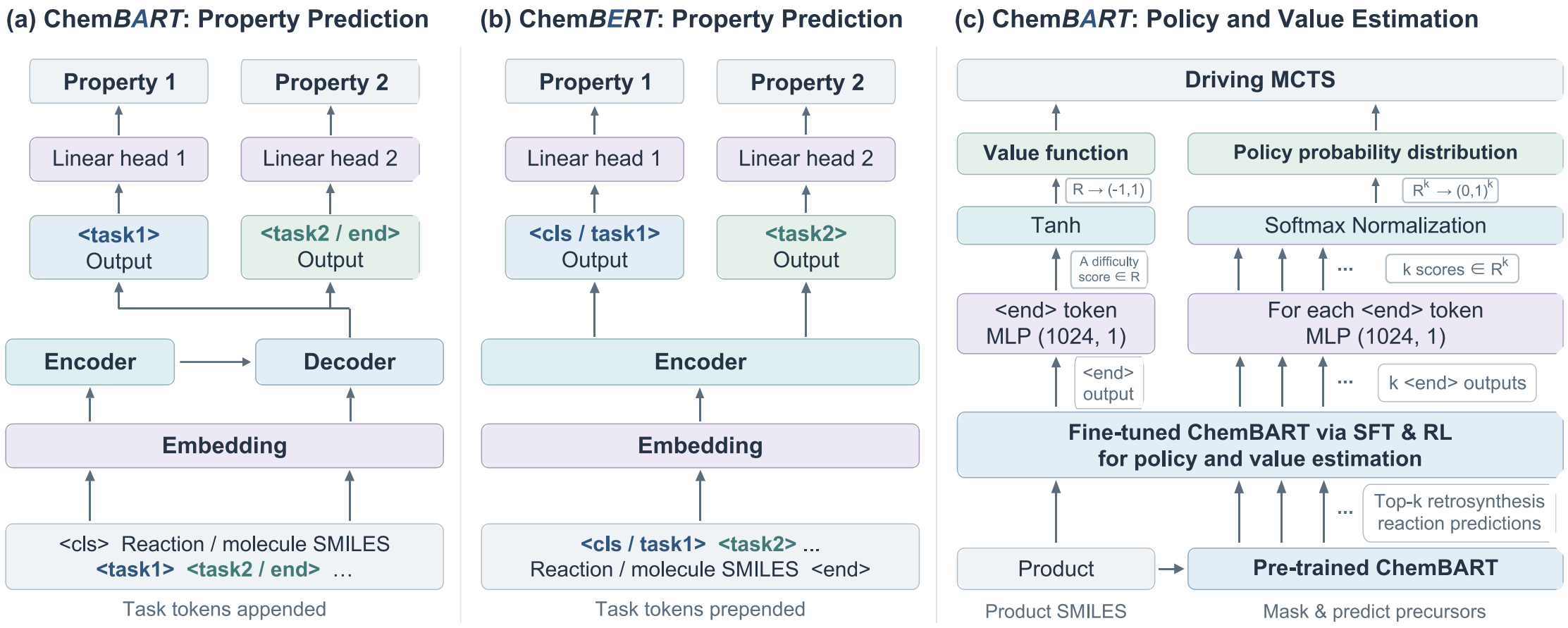

**Fig. 6 | Frameworks for classification, regression, and generation tasks using Transformers.** (a) BART's classification/regression scheme: Task-specific tokens are appended to the input sequence. The complete sequence is processed by both the encoder and decoder. The decoder's output corresponding to each task token is passed through a linear layer to obtain the classification or regression label. (b) BERT's classification/regression scheme: Task-specific tokens are prepended to the input sequence. The encoder processes the entire sequence, and the output corresponding to each task token is passed through a linear layer to obtain the classification or regression label. (c) Fitting policy and value functions: A pre-trained model generates multiple possible precursors for a specific product. The value function $v$ is obtained by inputting the product's expression into a fine-tuned regression model. The policy function $p$ is

derived by inputting each corresponding synthetic reaction for all precursor options into a fine-tuned regression model, then normalizing the output probabilities.

**Biochemical molecular property classification.** In this task, ChemBART is fine-tuned on five benchmark datasets: BBBP (bioactivity, 2,053), HIV (antiviral, 41,127), BACE (binding affinity, 1,514), Tox21 (toxicology, 8,014), and ClinTox (clinical toxicity, 1,484)[51], all structured as 8:1:1 training/validation/test splits.

We also explore the effectiveness of Low-Rank Adaptation (LoRA)[52] in fine-tuning on these datasets. LoRA offers the advantages of faster training and reduced memory consumption:

$$W' = W + \alpha BA$$

Where $W$ is the weight matrix of attention, frozen in LoRA fine-tuning, and $A \in R^{r*d}, B \in R^{d*r}$ are very small trainable weight matrices. For each task, two hyperparameters were tuned: the rank $\boldsymbol{r}$ governs the model's capacity, and the scaling factor $\boldsymbol{\alpha}$ controls the importance of the low-rank updates applied to the original weights.

We applied LoRA fine-tuning using the PEFT library from Hugging Face[101]. For the ClinTox dataset, we set 6 for both parameters, and for the other tasks, we used 8 for both, following hyperparameter optimization.

**MCTS Policy and Value Regression.** In the search tree of MCTS in our problem, a node is a single molecule with a value function. A node has multiple child nodes, which are choices of possible synthetic precursors. These child nodes have a probability distribution of being chosen, called the policy. Traditional multi-step synthesis approach like RetrosynthesisZero (ReSynZ)[29], Retro*[65], Retro*+[74], PDVN[66], EG-MCTS[67] and MEEA*[75] to fitting the policy and value function in MCTS was to use ECFP algorithm[102] to transform a molecular structure into a vector, and then use fully connected layers to yield policy and value with mlp_retrosyn library[68].

We propose a novel method, utilizing our pre-trained ChemBART model, to effectively fit the policy and value function in MCTS. We input the SMILES of the molecule at a node into ChemBART to fit the value function.

$$\mathrm{v(n)} = v_\Theta(\mathrm{m})$$

where n denotes a node, m denotes the SMILES expression of the molecule at the node, and Θ denotes the fine-tuned model parameters.

We then input each child node with its corresponding reagents and parent node into ChemBART to obtain a success score for the reaction. Next, we normalize the scores across all child nodes to derive the policy function.

$$\mathrm{p(n)} = [p_1, \dots p_n], \text{ where } p_i = p_\Theta(r_i)$$

$$\mathrm{p(n)} = \frac{p}{\sum_{i=1}^{n} p_i}$$

where $r_i$ denotes the SMILES expression of the $i^{th}$ candidate retrosynthesis reaction.

The algorithm is depicted in **Fig. 6c**. Compared with the traditional regression method of MCTS parameters, the embedding of Transformers serves a similar functionality to ECFP, converting molecule structure to a digital representation, while the encoder and decoder of Transformers are also similar to the fully connected layers, extracting high-level information of molecular structures. Our experimental results have proved the effectiveness of our novel model structure on this task.

Since running MCTS with LLMs is computationally expensive, requiring hundreds of LLM calls per instance, we did not run the reinforcement learning algorithm of ReSynZ directly on ChemBART through iterative MCTS execution and retraining. Instead, we adopted a cold-start approach by leveraging the ReSynZ program that had finished reinforcement learning to generate MCTS data efficiently, with 6,000 data points from successful retrosynthesis routes and 6,000 from failed ones. A data point is in the form of tuple [product, product's value, precursor candidate list, candidates' policy distribution]. For details of training data collection, please refer to Supplementary **Table S11**. The dataset was split into training, validation, and test sets in a 90:5:5 ratio. We then trained a template-based neural network from scratch, as well as ChemBART-F, -M and -R, respectively, using this dataset to compare performance. Subsequent reinforcement learning on ChemBART with MCTS yielded only marginal improvement in success rate (~2%), indicating that the model had already absorbed most of the relevant knowledge from the ReSynZ-generated data.

**Single Step Generation.** Our pre-training approach involves masking one of the three components, precursors, reagents, or products, within the reaction SMILES expression, prompting the model to output the complete reaction. This setup inherently allows for various predictive tasks, such as deriving precursors from products, inferring reagents based on both reactants and products, and predicting products given reactants and reagents, without requiring additional fine-tuning. We employed a beam search strategy for LLMs to generate k alternative answers for each input across these three tasks. In beam search, at each round of predicting the following tokens, only the generated sequences with top-$k$ cumulative probability are retained.

**Synthesis Route Planning with MCTS.** The MCTS[28] algorithm plays a key role in efficiently exploring potential synthetic pathways in large chemical reaction spaces. Each node in the search tree, representing a molecule, is initialized with two parameters: policy $p$ and value $v$, from neural networks. Each round of exploration tries to find a synthesis route, with each step decided by the parameters of the nodes. Through several times of exploration, the policy and value of each node would be gradually updated and improved to point to the optimal route. Finally, MCTS outputs the probability distribution $\boldsymbol{\Pi}$ of all the child nodes of the root node, i.e. the target molecule, by the following formula

$$\Pi = \frac{[N_1^\tau, N_2^\tau, \ldots]}{N^\tau}$$

where $\boldsymbol{N_i}$ is the number of visits of the $i^{th}$ child node, $\boldsymbol{N}$ is the total number of visits conducted, and $\boldsymbol{\tau}$ is the temperature coefficient. The $\boldsymbol{\Pi}$ can be seen as an improved policy $p$ of the root. If only the best route is needed, in each step the precursors with the highest probability in $\boldsymbol{\Pi}$ are greedily picked. For alternative synthesis routes for a target molecule, we also adopted beam search to pick the top-$k$ routes in each step.

To drive MCTS, the algorithms to generate all the child nodes, as well as the policy and value for each node are required.

**Generating candidate child nodes for a parent node.** We generated top-$k$ possible synthetic precursor candidates with three alternative sampling strategies. In addition to beam search, top-$k$ strategy retains top-$k$ tokens, while top-$p$ retains tokens with probabilities no less than p when generating each token. (Supplementary **Fig. S36**) To ensure the quality of the overall synthesis pathway, we filtered out unreliable predictions at each step. Using the RDKit[103] module, we canonicalized the SMILES of all generated child nodes and discard any that cannot be canonicalized, which indicates structurally invalid predictions. We then deduplicated child nodes that share the same canonical SMILES, summing their associated probabilities. Next, we used ChemBART to predict the reagents for each candidate retrosynthesis reaction, which are essential implementation details, and subsequently predict the product from each precursor-reagent pair. If the predicted product does not match the original parent node, the predicted reaction is considered incorrect and the corresponding child node is discarded. Finally, we normalized the probabilities of the remaining child nodes to get a probability distribution $\boldsymbol{p_1}$.

**Predicting value and policy functions.** We obtained the value function by inputting the target molecule at a node into our fine-tuned model. The policy $\boldsymbol{p_2}$ is derived by inputting each filtered candidate reaction at that node into the model to obtain a score, then normalized. However, the resulting policy is derived purely from reinforcement learning, which optimizes only the likelihood of successfully completing a synthesis pathway, without accounting for the practical feasibility of the individual reactions, which can only come from literature. To address this, we incorporated the generation probability $\boldsymbol{p_1}$ reflecting confidence based on literature, with the policy from the fine-tuned ChemBART model reflecting the ease of achieving complete synthesis, by multiplying them elementwise and then normalizing the resulting probability to yield the final policy $\boldsymbol{p^*}$.

$$p^* = (p_1 \odot p_2)$$

$$p^* = \frac{p^*}{\sum_{i=1}^{n} p_i^*}$$

It should also be noted that the methodology of combining the knowledge from literature and exploration is also the popular trend in recent CASP works like PDVN[66], Retro*+[74], EG-MCTS[67] and so on.

To optimize computing efficiency, a separate process is assigned to each target molecule while planning synthesis pathways, while ChemBART parameters (pre-trained and fine-tuned) on GPUs are shared across these processes.

After a complete synthesis route is designed, we used other fine-tuned ChemBART models stated above to add additional information to each step of the route, such as reaction temperature, reaction yield and molecular property of the molecules involved.

## Acknowledgements

This work was supported by the National Key R&D Program of China (No. 2023YB3813001), the National Natural Science Foundation of China (Nos. 22273034, 22361142831, 22193073 and 92253305), Jiangsu Province Front-end Technology Research and Development Program (BF2024056), the Fundamental Research Funds for the Central Universities (021514380026), the Beijing National Laboratory for Molecular Sciences (BNLMS-CXX-202106 to X.L.), and Nanjing University Tengya Innovation Fund. X. L. is supported by the New Cornerstone Science Foundation through the XPLORER PRIZE. J. W. is supported by the Boya Postdoctoral Fellowship of Peking University and the China Postdoctoral Science Foundation (GZB20250264). Parts of the calculations were performed using computational resources on an IBM Blade cluster system from the High-Performance Computing Center (HPCC) of Nanjing University.

## Author Contributions

K.N.L and H.D. conceptualized the work and designed the methodology. K.N.L, Y.J.Z., and H.P.G. performed the computational studies, J.W., Z.Y.S., and X.G.L. performed wet-lab validations. All authors discussed the data and contributed to the drafting. X.G.L. and H.D. supervised the project and acquired funding.

## Competing Interests

The authors declare no competing financial interests.

# Additional information

**Correspondence and requests for materials** should be addressed to Hao Dong.

# Data Availability

All the data used in this paper are from open-source datasets. Please check the reference in the paper for all of them. Source data of all the experiments have been made public in the code repository:
https://github.com/njukenanli/ChemBart_Planner/tree/main/paper_original_data.

The model weights would be available on Huggingface upon paper acceptance:
https://huggingface.co/ChemBart.

# Code Availability

The source code is available on GitHub:
https://github.com/njukenanli/ChemBART_Planner.

# Supplementary Information

Please refer to:
https://drive.google.com/file/d/1zCmbRNnhuWeeXanhHDsWg5ySleU8XclS/view?usp=sharing.